\pdfoutput=1

\documentclass[11pt]{article}

\usepackage{ACL2023}

\usepackage{times}
\usepackage{latexsym}

\usepackage[T1]{fontenc}

\usepackage[utf8]{inputenc}

\usepackage{inconsolata}
\usepackage{microtype}
\usepackage{graphicx}
\usepackage{amsmath}
\usepackage{amssymb}
\usepackage{booktabs}

\usepackage{multirow}
\usepackage{appendix}
\usepackage{dutchcal}
\usepackage{bm}
\usepackage{booktabs,makecell,multirow}
\usepackage{xcolor}

\title{Modularized Zero-shot VQA with Pre-trained Models}
\author{Rui Cao \and Jing Jiang \\
  School of Computing and Information Systems \\
  Singapore Management University \\
  \texttt{ruicao.2020@phdcs.smu.edu.sg}, \texttt{jingjiang@smu.edu.sg} \\}

\begin{document}
\maketitle

\begin{abstract}
Large-scale pre-trained models~(PTMs) show great zero-shot capabilities.
In this paper, we study how to leverage them for zero-shot visual question answering~(VQA).
Our approach is motivated by a few observations.
First, VQA questions often require multiple steps of reasoning, which is still a capability that most PTMs lack.
Second, different steps in VQA reasoning chains require different skills such as object detection and relational reasoning, but a single PTM may not possess all these skills.
Third, recent work on zero-shot VQA does not explicitly consider multi-step reasoning chains, which makes them less interpretable compared with a decomposition-based approach.
We propose a modularized zero-shot network that explicitly decomposes questions into sub reasoning steps and is highly interpretable. 
We convert sub reasoning tasks to acceptable objectives of PTMs and assign tasks to proper PTMs without any adaptation.
Our experiments on two VQA benchmarks under the zero-shot setting demonstrate the effectiveness of our method and better interpretability compared with several baselines.
\end{abstract}

\section{Introduction}
\label{sec:intro}

\begin{figure*}[ht!] 
	\centering
	\setlength{\tabcolsep}{0pt} 
	\renewcommand{\arraystretch}{0} 
	\begin{tabular}{ccc}
		\includegraphics[scale = 0.27]{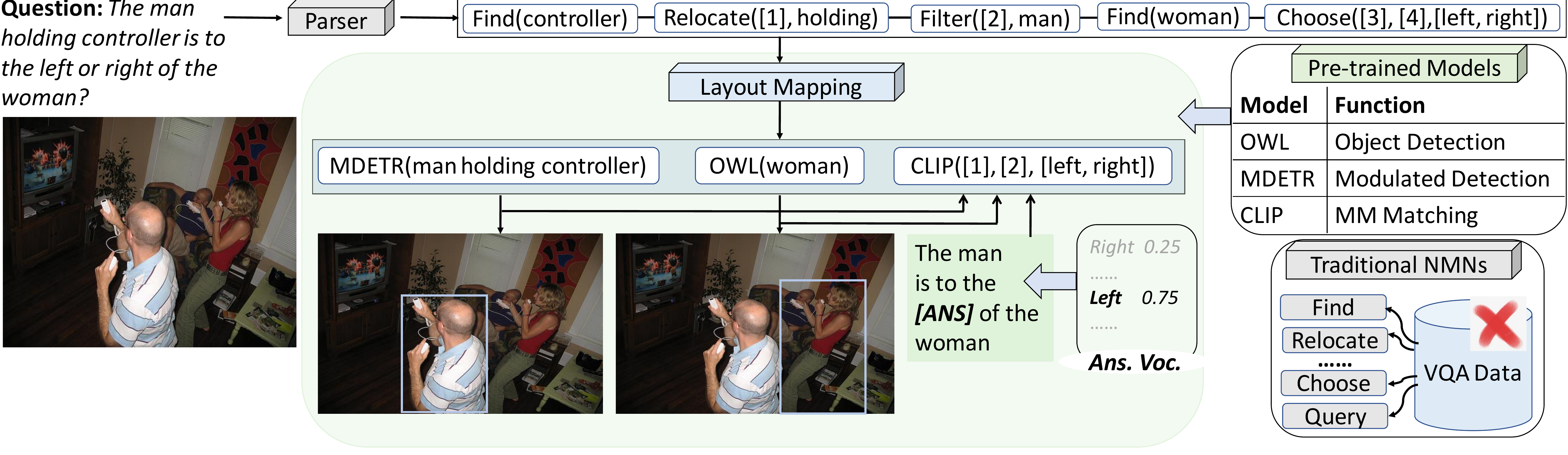} 
	\end{tabular}
	\caption{An overview of our proposed method. Instead of training modules in NMN, we propose a modularized zero-shot VQA method leveraging pre-trained models to perform different reasoning tasks.}
	\label{fig:arch}
\end{figure*}

Visual Question Answering~(VQA), the task of answering textual queries based on information contained in an image, is a multimodal task that requires comprehension and reasoning of both visual and textual content~\cite{DBLP:journals/ijcv/AgrawalLAMZPB17,DBLP:conf/cvpr/HudsonM19}. 
Most previous work on VQA either trains VQA models from scratch~(e.g.,~\citet{DBLP:conf/emnlp/FukuiPYRDR16,DBLP:conf/cvpr/00010BT0GZ18}) or fine-tunes pre-trained vision-language models for VQA~(e.g.,~\citet{DBLP:journals/corr/abs-1908-03557,DBLP:conf/nips/LuBPL19}).
Thus, they rely heavily on labeled VQA data, which are expensive to obtain. 
VQA models based on supervised learning are also hard to generalize to new domains or new datasets~\cite{DBLP:conf/emnlp/XuCCDL20,DBLP:conf/cvpr/ChaoHS18,DBLP:journals/corr/abs-2103-15974}.

Recently, large-scale pre-trained models (PTMs) have demonstrated strong transferability to different downstream tasks under zero-shot settings, i.e., without any training data for the downstream tasks~\cite{DBLP:conf/nips/BrownMRSKDNSSAA20,DBLP:conf/icml/RadfordKHRGASAM21}.
With increased pre-training data size, these models show strong zero-shot performance on various down-stream tasks, such as image classification and face detection with the CLIP model~\cite{DBLP:conf/icml/RadfordKHRGASAM21} and sentiment analysis and commonsense question answering with the GPT-3 model~\cite{DBLP:conf/nips/BrownMRSKDNSSAA20}. 
However, few studies have focused on zero-shot VQA from pre-trained models.

Despite the power of these PTMs, it is not straightforward to directly apply them to VQA under zero-shot settings, because they are not pre-trained with the same objective as VQA. 
Some recent work converts images to tokens that pre-trained language models can understand so that VQA can be converted to text-based QA~\cite{DBLP:conf/aaai/YangGW0L0W22,DBLP:journals/corr/abs-2210-08773,DBLP:conf/nips/TsimpoukelliMCE21,DBLP:conf/acl/Jin0SC022,DBLP:conf/acl/DaiHSJLF22}. 
However, this approach requires either a strong pre-trained image captioning model that can capture sufficient visual details or auxiliary training to obtain such a captioning model.
Some other work converts VQA into a multimodal matching problem so that pre-trained vision-language models~(PT-VLMs) such as CLIP can be used~\cite{DBLP:conf/acl/0002000W22,DBLP:conf/iclr/ShenLTBRCYK22}. 
However, complex VQA questions such as those found in the GQA dataset~\cite{DBLP:conf/cvpr/HudsonM19} often require spatial reasoning
and/or multi-step reasoning, 
which PT-VLMs may not be strong at~\cite{DBLP:conf/acl/SubramanianMD0022, DBLP:journals/corr/abs-2204-03162}.

VQA questions can be complicated and often require different reasoning steps such as object detection and spatial reasoning, as the example question in Figure~\ref{fig:arch} illustrates.
Previously, people proposed Neural Module Networks~\cite{DBLP:conf/cvpr/AndreasRDK16, DBLP:conf/iccv/HuARDS17}, which are modularized networks where each pre-defined module performs a specific reasoning task. 
These pre-defined modules are trained end-to-end from labeled VQA data. 
Motivated by the idea of modularization, in this paper, we propose a modularized zero-shot network for VQA~(\textbf{Mod-Zero-VQA}) by decomposing questions into sub-tasks and assigning appropriate sub-tasks to PTMs without any adaptation.
Given a question, we first parse the question into basic reasoning steps explicitly. 
These reasoning steps will then be reconfigured and mapped to different PTMs based on a set of rules we define.
Specifically, we consider the following PTMs: \textit{OWL}~\cite{DBLP:journals/corr/abs-2205-06230} as the object detector, \textit{MDETR}~\cite{DBLP:conf/iccv/KamathSLSMC21} for reference expression localization (including several skills such as relational and spatial reasoning) and \textit{CLIP}~\cite{DBLP:conf/icml/RadfordKHRGASAM21} as the answer generator for open-ended questions. 
Considering the limited capabilities of current pre-trained vision-language models in spatial relation understanding~\cite{DBLP:conf/acl/SubramanianMD0022}, we also define simple and general heuristics to aid spatial reasoning. 
Note that only when we decompose questions and reasoning chains step by step can we insert human heuristics for spatial reasoning, because we have the intermediate outputs such as objects' bounding boxes from previous steps. 

We evaluate the proposed method on the GQA dataset~\cite{DBLP:conf/cvpr/HudsonM19} where questions are compositional and require multi-step reasoning. 
The experiment result shows that the proposed model surpasses the baselines significantly on GQA, with near $\mathbf{13\%}$ of relative improvement over the strongest baseline (from 41.9 to 47.3).
The results confirm the benefit of modualarization when using PTMs for zero-shot VQA.
In addition, our method is interpretable because of the explicit reasoning steps generated.

The contributions of our work can be summarized as follows: 
(1) We propose a novel modularized zero-shot VQA method that utilizes different pre-trained models for different reasoning steps; (2) We design rules to map different VQA reasoning steps to suitable PTMs so that we can leverage these PTMs without any adaptation; 3) Experiment results show the effectiveness of the proposed method, especially when questions consist of multiple steps of reasoning.

\section{Background}
\label{sec:background}


\paragraph{Task Definition.}
Given an image $I$ and a question $Q$, a VQA system is expected to return an answer $a$.
Traditional fully supervised VQA relies on a training set consisting of (image, question, answer) triplets.
For zero-shot VQA, no such training data is given.
However, in this paper we assume that we can use pre-trained models (PTMs) to help us with zero-shot VQA.

\paragraph{Existing Zero-shot VQA Methods.}
Work on zero-shot VQA is very limited.
We can organize existing work into the following categories.
One line of work leverages the question answering capability in pre-trained language model (LMs). 
Some of them adopt prefix language modeling with weakly-supervised data other than VQA data (i.e., image-text pairs) to convert visual information into discrete tokens (prefix) that LMs can understand.
Frozen~\cite{DBLP:conf/nips/TsimpoukelliMCE21}, VLKD~\cite{DBLP:conf/acl/DaiHSJLF22} and FewVLM~\cite{DBLP:conf/acl/Jin0SC022} fall under this category.
Some directly convert VQA images into textual descriptions so that the task of VQA changes to text-based QA and LMs can be applied.
Methods in this category include PICa~\cite{DBLP:conf/aaai/YangGW0L0W22} and PnP-VQA~\cite{DBLP:journals/corr/abs-2210-08773}.
Recent work~\cite{DBLP:conf/acl/0002000W22,DBLP:conf/iclr/ShenLTBRCYK22} converts VQA to an image-text matching problem and prompts the CLIP model~\cite{DBLP:conf/icml/RadfordKHRGASAM21}, a large-scale vision-language model pre-trained on the image-text matching task. 
The prompts can be either question irrelevant such as \textit{Quesion: [\texttt{Ques}]; Answer: [\texttt{MASK}]}~(QIP by \citet{DBLP:conf/iclr/ShenLTBRCYK22}) or question-related by converting questions into a masked statement~(TAC-P by \citet{DBLP:conf/acl/0002000W22}).

However, 
a limitation with these methods is that several of them still require training, although the training data is not in the form of VQA.
Besides, converting images to captions and leveraging text-based QA may lose important visual details during the caption generation step.
The two methods above using CLIP do not address the issue that CLIP model lacks compositional and  spatial reasoning abilities, which has been observed in previous work~\cite{DBLP:conf/acl/SubramanianMD0022, DBLP:journals/corr/abs-2204-03162}.




\section{Modularized Zero-shot VQA}
\label{sec:model}

Our method is motivated by Neural Module Network~(NMN) based VQA, which decomposes questions into reasoning steps, where each module in the NMN is pre-defined to perform a specific reasoning task. 
The idea allows us to select appropriate pre-trained models to handle different reasoning tasks in a question.
Specifically, in NMN-based VQA, we first manually define a set of reasoning steps such as object detection and spatial reasoning, each represented by a \textit{module}.
A question is then explicitly decomposed and converted into a \textit{layout} of modules, which is an executable program showing the reasoning chain to reach the final answer.
The top section of Figure~\ref{fig:arch} shows the layout corresponding to the sample question.
To train an NMN-based VQA system, usually a layout generator is separately built first, which either uses hand-crafted rules over dependency parses of questions or is a trained seq2seq model.
Then, the parameters of the various VQA modules are learned from VQA training data.

For our work, we do not want to use VQA data for training.
But we observe that many modules in NMN-based VQA can be supported by pre-trained models that have already acquired the capabilities needed by these modules.
The key component of our method is therefore to map a layout of modules produced by traditional NMN-based VQA to a simplified layout of zero-shot components that can be implemented directly using pre-trained models.

%

\subsection{Traditional VQA Modules}
\label{subsec:traditional_modules}
There is not any standard set of modules for VQA.
We largely adopt the design of modules introduced by \citet{DBLP:conf/iccv/HuARDS17} with some minor changes.
We assume that the image has been pre-processed and $N$ bounding boxes have been detected, each represented as an embedding vector, collectively denoted as $\mathbf{V} = (\mathbf{v}_1, \mathbf{v}_2, \ldots, \mathbf{v}_N)$.
An attention map $\bm{\alpha}$ is defined to be a distribution over the $N$ bounding boxes.


Table~\ref{tab:subset_of_modules} lists the most important traditional VQA modules that we will replace with pre-trained models.
The full list of modules can be found in Table~\ref{tab:nmn-module} in the appendices.
It is worth explaining that besides taking in $\mathbf{V}$ and $\bm{\alpha}$ as either input or output, many modules also take in the word embeddings of some text description extracted from the question.
These text embeddings are arguments to control the behaviors of the modules.
For example, the $\texttt{Find}$ module's objective is to locate an object among all the bounding boxes given.
The textual input $\bf{g}_\text{OBJ}$ is therefore the word embedding of the name of the object to be found.
Similarly, $\bf{g}_\text{RELA}$ ,$\mathbf{g}_{\text{ATTR}}$ and $\mathbf{g}_{\text{QUERY}}$ are word embeddings for the description of relation (e.g., \textit{to the left of}), attribute (e.g., \textit{red}) and aspect to query (e.g., querying \textit{name}).

\begin{table}[ht]
  \begin{tabular}{ll} 
    \toprule
    \textbf{Module} &  \textbf{Inputs}\\
    \midrule
    \texttt{Find} & $\mathbf{V}$, $\mathbf{g}_{\text{OBJ}}$\\
    \texttt{Relocate} & $\bm{\alpha}$, $\mathbf{V}$, $\mathbf{g}_{\text{RELA}}$\\
     \texttt{Filter}&$\bm{\alpha}$, $\mathbf{V}$, $\mathbf{g}_{\text{CONDI}}$\\
     \midrule
    \texttt{Choose}& $\bm{\alpha}_1$, $\bm{\alpha}_2$, $\mathbf{V}$, $\mathbf{g}_{\text{RELA}^1}$, $\mathbf{g}_{\text{RELA}^2}$\\
    \texttt{Query} & $\bm{\alpha}$, $\mathbf{V}$, $\mathbf{g}_{\text{QUERY}}$\\
    \bottomrule
\end{tabular}
\centering
\caption{A subset of the modules in traditional NMN that we replace with pre-trained models. Modules in the first block output an attention map and those in the second block generate an answer.}
\label{tab:subset_of_modules}
\end{table}

Traditionally, the parameters of the modules in Table~\ref{tab:subset_of_modules} need to be learned from VQA training data.
In other words, these modules' underlying capabilities such as object recognition and relational reasoning need to be acquired from VQA data.
However, we hypothesize that recently developed pre-trained models may already have some of these capabilities and can therefore directly equip these modules with such capabilities.
For example, the \texttt{Find} module is mainly responsible for object recognition,
and previously the parameters of \texttt{Find} have to be learned from scratch using VQA data.
Now with a powerful pre-trained model such as OWL~\cite{DBLP:journals/corr/abs-2205-06230} that can recognize a wide range of objects, we can presumably directly use a model like OWL to replace the traditional \texttt{Find} module.

\subsection{Pre-trained Models}
\label{subsec:ptms}


We utilize three pre-trained models that we believe are highly relevant to VQA.

\paragraph{OWL.} The Vision Transformer for Open-World Localization~(OWL) model~\cite{DBLP:journals/corr/abs-2205-06230} is a model for open-vocabulary object detection.
It is first pre-trained on large-scale image-text pairs and then fine-tuned with added detection heads and medium-sized detection data.
Given the category name of an object and an image, the model is able to locate bounding box(es) in the image containing the object together with a confidence score for each box.

\paragraph{MDETR.} The modulated DETER~(DEtection TRansformer) model~\cite{DBLP:conf/iccv/KamathSLSMC21} is an end-to-end detector that can detect an object in an image conditioned on a piece of textual description of the object such as its attributes and its relation with another object in the image.
The model is pre-trained on image-text pairs with explicit alignment between phrases in the text and bounding boxes of objects in the image.
Given an image and the description of an object, MDETR is able to locate the bounding box(es) in the image containing the object satisfying the description.
Note that different from OWL, MDETR is able to understand textual descriptions that may contain attribute information and/or complex visual relations.
For example, given the description \textit{a man holding a yellow cup is talking}, MDETR will detect the bounding box containing the man holding a yellow cup in the given image, whereas OWL is not able to use the description and will only recognize all bounding boxes containing a man. Note that we use the version of MDETR pre-trained on general modulated detection \textbf{without fine-tuning} for any downstream tasks.

\paragraph{CLIP.} CLIP is a well-known large-scale vision-language model by OpenAI.
It is pre-trained with 400M image-caption pairs through contrastive learning.
Given an (image, text) pair, CLIP uses its separate image encoder and text encoder to turn the image and the text each into a vector, and the cosine similarity between the two vectors directly measures the compatibility of the two.
Recent work has shown that CLIP can be directly used for VQA in a zero-shot setting, if we can come up with a set of candidate answers and transform each (question, answer) pair into a statement~\cite{DBLP:conf/acl/0002000W22}.

\subsection{Zero-shot NMN using Pre-trained Models}
\label{subsec:mapping}

Based on the descriptions of the traditional VQA modules in Section~\ref{subsec:traditional_modules} and of the three PTMs we consider in Section~\ref{subsec:ptms}, we can see that there are obvious connections between the capabilities desired by the traditional modules and the capabilities that these PTMs have already acquired.

However, the mapping between them is not trivial.
First of all, there is no simple one-to-one mapping from traditional VQA modules to the PTMs.
For example, the MDETR model can already perform multiple steps of reasoning to locate the desired object, so it can be used to cover a sequence of modules in an NMN layout.
Second, there may be capabilities required when applying PTMs but not captured by modules defined in NMN-based VQA.
In particular, the MDETR model always assumes that the object to be grounded exists in the given image, but for those questions asking for the existence of a specified object, we cannot directly use MDETR.

To address these challenges, we carefully design a mapping mechanism that can map an NMN-based module layout to a simplified layout consisting of a few zero-shot modules.
Three of these zero-shot modules (\texttt{OWL}, \texttt{MDETR} and \texttt{CLIP}) correspond exactly to the three PTMs introduced earlier.
The rest of the zero-shot modules are defined by simple heuristic rules.
We list these zero-shot modules in Table~\ref{tab:zero_modules}.
\begin{table}[ht]
  \begin{tabular}{lll} 
    \toprule
    \textbf{Module} &  \textbf{Inputs}&  \textbf{Output}\\
    \midrule
    \texttt{OWL} & $I$, OBJ & $\mathcal{B}$, $\mathbf{s}$\\
    \texttt{MDETR} & $I$, SENT & $\mathcal{B}$, $\mathbf{s}$\\
    \texttt{CLIP}& $\mathcal{B}$, $I$, $\mathcal{V}$  & Ans.\\
    \midrule
    \texttt{Count} & $\mathcal{B}$ & Num.\\
    \texttt{Exist} & $\mathcal{B}$, (ATTR/RELA) &\textit{Yes/No}\\
    \texttt{And} &\texttt{Exist}$_1$, \texttt{Exist}$_2$& \textit{Yes/No}\\
    \texttt{Or} &\texttt{Exist}$_1$, \texttt{Exist}$_2$& \textit{Yes/No}\\
    \bottomrule
\end{tabular}
\centering
\caption{Zero-shot modules with either pre-trained models or heuristics. The $I$ is the VQA image, $\mathcal{V}$ is the answer vocabulary and $\mathcal{B}$ is the set of bounding boxes.}
\label{tab:zero_modules}
\end{table}

We now give a high-level summary of the mapping mechanism below.
We first look at the last module in the NMN layout.
If the last module is one of \texttt{Choose}, \texttt{Compare} and \texttt{Query}, we know that the input to this last module is either a single attention map or two attention maps, where each attention map essentially tries to capture an object matching some textual descriptions.
By tracing the path in the layout leading to the attention map, we choose either the zero-shot \texttt{OWL} module (when the path has a length of 1) or the zero-shot \texttt{MDETR} module (when the path is longer than 1 hop).
This is because when the path length equals to one, it involves only object detection (corresponding to a single \texttt{Find} module in the NMN layout for generation of the attention map). 
When the path length is more than one, it indicates the generation of the attention map in the NMN layout involves other modules such as  \texttt{Filter} and \texttt{Relocate}, which calls for the other abilities than object detection, such as language understanding, attribute recognition and relational reasoning. 
Different from NMN modules which takes in image features and object embeddings to generate an attention map, our zero-shot \texttt{OWL} and zero-shot \texttt{MDETR}  takes in the raw image and raw texts to locate (OBJ for \texttt{OWL} and SENT for \texttt{MDETR}) to generate a set of detected bounding boxes $\mathcal{B}=\{\mathbf{b}_n\}_{n=1}^N$ together with their confident scores $\mathbf{s} \in \mathbb{R}^N$, where $\mathbf{b}_n \in \mathbb{R}^4$ represents the relative position and size of the detected bounding box in the image.
We keep only the bounding box from either \texttt{OWL} or \texttt{MDETR} with the highest confident score and feed it to  \texttt{CLIP}. 
We generate an answer by leveraging the capability of multimodal matching of CLIP. 
Specifically, given $\mathcal{B}$, we generate an input image (which we refer to as $I^{\text{in}}$) by either masking regions not containing those detected boxes ($|\mathcal{B}|=2$) or cropping the image so that only the part containing the box remains ($|\mathcal{B}|=1$). If the final NMN module is \texttt{Choose}, we generate a masked template by question conversion as in~\cite{DBLP:conf/acl/0002000W22}; otherwise the masked template will be a simple ``[\texttt{MASK}]''. 
Then we match the image $I^{\text{in}}$ 
with the template where the [\texttt{MASK}] token is replaced by each of the answer candidates in $\mathcal{V}$. 
We then select the answer that, when placed inside the template, best matches the image.

If the module is \texttt{Exist}, we trace back the path leading to \texttt{Exist} to determine whether the module is asking for the existence of an object, an attribute or a relation.
For object existence (e.g., \textit{is there a car}), 
we use the zero-shot \texttt{OWL} module.
For attribute existence and relation existence, we first verify whether all mentioned nouns (objects) detected by a POS tagger in the question exist with the \texttt{OWL} module. 
Once we detect an object that does not exist, the predicted answer will be \textit{no}.
If all objects exist, then we generate corresponding bounding boxes 
leveraging either \texttt{OWL} or \texttt{MDETR} following the method described in the paragraph above. 
For attribute existence, we generate a pair of a positive and a negative descriptions: (ATTR, \textit{not }ATTR), e.g., (\textit{red}, \textit{not red}).
We then find which description aligns better with the cropped image according to $\mathbf{b}$. 
If the image aligns better with the positive statement, then the answer will be  \textit{yes}; otherwise, \textit{no}. 
For relation existence, we generate the masked image $I^{\text{in}}$ 
according to $\mathbf{b}_1$ and $\mathbf{b}_2$ (the bounding boxes of the two objects whose relation is to be checked) 
and a pair of opposite statements regarding the relation to be checked, following~\cite{DBLP:conf/acl/0002000W22}.
For example, if the question is to check whether \textit{A} is holding \textit{B}, the two opposite statements will be \textit{A is holding B} and \textit{A is not holding B}.
For both attribute and relation existence, we use zero-shot \texttt{CLIP} for the alignment between the input image and the statements.
More details and the work flows of existence-related questions are provided in Appendix~\ref{sec:existence-ques}.

If the module is \texttt{Count}, we directly count the number of bounding boxes in $\mathcal{B}$ returned either from \texttt{OWL} or \texttt{MDETR}.
Finally, if the last module is a logical \texttt{AND} or logical \texttt{OR}, we further trace to the inputs of this module, which should both be an \texttt{Exist} module.
We then use the same mechanism described above for \texttt{Exist} to process the module. By receiving the outputs from the \texttt{Exist} modules, logical operations will be applied to determine the output. The deterministic logical operations can be found in Appendix~\ref{sec:logical-op}.




\subsection{Spatial Heuristics}
As mentioned in~\cite{DBLP:conf/acl/SubramanianMD0022}, CLIP is less capable of spatial reasoning. 
Using CLIP for answer generation may not be enough when it involves spatial relation understanding.
Following~\cite{DBLP:conf/acl/SubramanianMD0022}, we define simple and general heuristics to perform certain types of spatial reasoning. Note that only when we decompose questions explicitly can we insert the spatial heuristics into CLIP-based answer generation because we have the intermediate outputs from previous reasoning steps.

First of all, given the coordinates and the size of a bounding box, we use manual rules (named as \textbf{SpD}) to decide its position in the image as \textit{left, right, bottom, top}. 
Besides, we define heuristics, denoted as \textbf{SpC}, to solve spatial relations between two bounding boxes
(e.g., \textit{to the left of} and \textit{to the right of}).

Details of the implementation of the spatial relation solvers can be found in Appendix~\ref{sec:sp-heur-detail}.

\section{Experiments}
\label{sec:exp}

\subsection{Dataset}
We evaluate the proposed modularized zero-shot VQA method on two benchmarks: GQA~\cite{DBLP:conf/cvpr/HudsonM19} and VQAv2~\cite{DBLP:conf/cvpr/GoyalKSBP17}. The GQA dataset consists of questions requiring multi-step reasoning and various reasoning skills.
Around $94\%$ of the questions require multiple reasoning steps. 
We regard it as the main dataset to demonstrate the effectiveness of the proposed method compared with the baselines.
Compared with GQA, questions on the VQAv2 dataset   require fewer reasoning steps and are of diverse semantics. We use VQAv2 to show the validity of our method in real-world VQA.
We report standard accuracy for the GQA dataset while soft accuracy~\cite{DBLP:conf/cvpr/GoyalKSBP17} for VQAv2 dataset as there are multiple ground-truth answers.
We report the statistics of the datasets in Appendix~\ref{sec:dataset-st}.

\subsection{Implementation Details}
We conduct experiments on NVIDIA Tesla V100 GPU. The thresholds for the OWL and the MDETR model to filter out detected bounding boxes of low confidente scores are set to be $0.2$ and $0.7$ respectively. 
We follow~\cite{DBLP:conf/acl/0002000W22} for the generation of the answer vocabulary $\mathcal{V}$ for open-ended questions.
More details about answer vocabulary generation can be found in Appendix~\ref{sec:ans-gen} and more information about experiment settings can be found in Appendix~\ref{sec:exp-setup}.

\subsection{Main Results}
Zero-shot VQA performance of the baselines mentioned in Section~\ref{sec:background} and our proposed method are summarized in Table~\ref{tab:exp-all}\footnote{For FEWVLM and PNP-VQA model, we show their reported performances on GQA test-dev, which should have similar distributions as the validation split of GQA.}.

\begin{table}[t]
\centering
\begin{small}
  \begin{tabular}{l|cc}
    \toprule
    \textbf{Method} & \textbf{GQA} & \textbf{VQA}\\
    \midrule
    Frozen&- &29.5\\
    VLKD$_{\text{ViT-L/14}}$ &- &42.6\\
    FEWVLM$_{\text{base}}$&27.0& 43.4 \\
    FEWVLM$_{\text{large}}$&29.3 &47.7 \\
    PNP-VQA$_{6\text{M}}$&34.6 &54.3 \\
    PNP-VQA$_{11\text{B}}$&\textbf{41.9} &\textbf{63.3} \\
    \midrule
    QIP &35.9&21.4 \\
    TAP-C&36.3&38.7 \\
    Mod-Zero-VQA &\textbf{47.3} &\textbf{41.0}\\
    \midrule
\end{tabular}
\end{small}
\caption{Experimental results on the GQA and VQA datasets. The first block are models using the text-based QA capability of LMs and the second blocks are models incorporating CLIP.}
  \label{tab:exp-all}
\end{table}

First of all, we observe that the proposed Mod-Zero-VQA method ismore effective on the GQA dataset, which contains many multi-step reasoning questions. 
Mod-Zero-VQA clearly surpasses all baselines on GQA. 
The results suggest that it is effective under zero-shot settings to decompose questions when questions are compositional and require several steps of reasoning to reach the answer. Such decomposition allows us to take advantage of the capabilities of different pre-trained models.
We also test the validity of the proposed method on real-world VQAv2 dataset, where questions require fewer reasoning steps and of diverse semantics. 
We can see that our method still achieves the best performance among zero-shot methods that utilize CLIP.
Although better performance is achieved by several methods that utilize large language models (as shown in the first block of Table~\ref{tab:exp-all}), it is worth pointing out that these methods often require caption generation as a pre-processing step, and this step poses challenges. 
For example,
PNP-VQA generates $\mathbf{100}$ captions per question, which is laborious. 
There may also be redundancy because many captions are irrelevant for question answering. 
Another advantage of our Mod-Zero-VQA method over the other zero-shot baselines is that our method offers high interpretability by showing the explicit multi-step reasoning chain, which has not been considered by any previous work.
With question decomposition, we can design modularized networks and assign reasoning tasks to pre-trained models (PTMs) which are more capable of the tasks, and with more powerful pre-trained models coming out, our method can be easily extended to utilize newer and more effective PTMs.
Meanwhile, it is easier to pinpoint the weakest chain in a system and insert human heuristics to aid these modules.

\subsection{Ablation Study}
\label{sec:ablation}
In our Mod-Zero-VQA method, PTMs play an important role. 
In this section, we show the performance of Mod-Zero-VQA when we replace PTMs listed in Section~\ref{subsec:ptms} with alternative models.

\begin{table}[t]
\centering
\begin{small}
  \begin{tabular}{l|ccc}
    \toprule
     \textbf{Detector}& \textbf{Yes/No Qns} & \textbf{Other Qns} & \textbf{Overall}\\
    \midrule
    CLIP-FR &56.80 &33.82 &41.39 \\
    OWL &69.26 & 36.48& 47.28\\
    GT &76.48 &38.06 &50.72 \\
    \midrule
\end{tabular}
\end{small}
\caption{Performance of Mod-Zero-VQA with different object detectors on GQA.}
  \label{tab:diff_det_gqa}
\end{table}
\begin{table}[t]
\centering
\begin{small}
  \begin{tabular}{ll|c}
    \toprule
    \textbf{Method} & \textbf{PT-VLMs} & \textbf{Overall}\\
    \midrule
    \multirow{3}{*}{QIP} 
    &$\text{CLIP}_{\text{ViT-B/16}}$  &35.93\\
    &$\text{CLIP}_{\text{Res}50\times 16}$  &35.11\\
    &ALBEF &34.75\\
    \midrule
    \multirow{3}{*}{TAP-C} 
    &$\text{CLIP}_{\text{ViT-B/16}}$  &36.32\\
    &$\text{CLIP}_{\text{Res}50\times 16}$&38.16\\
    &ALBEF &38.36\\
    \midrule
    \multirow{3}{*}{Mod-Zero-VQA} 
    &$\text{CLIP}_{\text{ViT-B/16}}$ & 47.28\\
    &$\text{CLIP}_{\text{Res}50\times 16}$ & 46.49\\
    &ALBEF&\textbf{48.68} \\
    \midrule
\end{tabular}
\end{small}
\caption{Performance of the Mod-Zero-VQA model with different PT-VLMs as the zero-shot \texttt{CLIP} for answer generation on GQA.}
  \label{tab:diff_vl_gqa}
\end{table}

\noindent\textbf{Replacing OWL:} We tried replacing OWL with other object detectors. 
First, we consider an object detector combining Faster-RCNN~\cite{DBLP:conf/nips/RenHGS15} and CLIP (\textbf{CLIP-FR}). 
Specifically, Faster-RCNN is used to detect objects in an image and CLIP is applied to classify each detected object.
Second, we use the ground-truth object annotations from Visual Genome~\cite{DBLP:journals/ijcv/KrishnaZGJHKCKL17} to replace object detection results (\textbf{GT}), which serves as an upper bound. 
Results of our zero-shot NMNs with different object detectors are provided in Table~\ref{tab:diff_det_gqa}. 
We divide the questions into Yes/No (bindary) questions and other questions.
We observe that the quality of object detection is important to the performance of zero-shot NMNs. 
Our model with OWL surpasses the one with CLIP-FR, which has poorer detection performance than OWL. 
We also observe more substantial performance drop with binary questions.
We believe that this is because these questions are mostly about the existence of objects, so the object detection results affect the VQA performance more. 
Using Mod-Zero-VQA with the ground-truth object detection results would further improve the performance, as shown in the last row of Table~\ref{tab:diff_det_gqa}.
This suggests that when more accurate object detection models are developed, we can further improve the zero-shot VQA performance with our approach.

\noindent\textbf{Replacing CLIP: } 
We show the performance of replacing zero-shot \texttt{CLIP} (which is $\text{CLIP}_{\text{ViT-B/16}}$ by default in our experiments), with either $\text{CLIP}_{\text{Res}50\times 16}$ or ALBEF~\cite{DBLP:conf/nips/LiSGJXH21}, in Table~\ref{tab:diff_vl_gqa}. Because QIP and TAC-P convert VQA to a multi-modal matching task and both use PT-VLMs as the answer generator, we also replace the original $\text{CLIP}_{\text{ViT-B/16}}$ in these two baselines with the other PTMs.
We observe that Mod-Zero-VQA gives stable performance regardless of the vision-language model used, and it always outperforms the baselines substantially. 
This indicates that these PTMs can all be good substitutes for the zero-shot \texttt{CLIP} module. 
Compared with the two CLIP models (i.e., with either ViT~\cite{DBLP:conf/iclr/DosovitskiyB0WZ21} or ResNet~\cite{DBLP:conf/cvpr/HeZRS16} as the visual backbone), we also notice that using ALBEF~\cite{DBLP:conf/nips/LiSGJXH21} as the answer generator can enhance the performance. 
To better understand the advantage of using ALBEF over CLIP, we provide more detailed performance in Table~\ref{tab:detailed_gqa} in Appendix~\ref{sec:detailed-results}.
ALBEF mostly benefits the proposed method in 
the \textit{Query} type of questions, which usually ask about \textit{objects}, \textit{attributes} and \textit{relations}.
Consistent with~\cite{DBLP:journals/corr/abs-2207-00221}, end-to-end models~(i.e., ALBEF in this case) perform better than dual-encoder models~(i.e., CLIP in this case) in vision understanding tasks on average. 
A future direction may be to select the best pre-trained model per question.

\subsection{Out-of-Domain Generalization}
\label{sec:ood-gen}

Because our Mod-Zero-VQA method is not trained on any domain-specific VQA data but rather utilizes pre-trained models that are supposedly trained on data from a wide range of domains, we suspect that our Mod-Zero-VQA method is more robust across different domains compared with VQA models trained on specific domains and applied in cross-domain settings. 
We therefore also compare our Mod-Zero-VQA with fully-supervised models in the Out-of-Domain Generalization (OOD) setting. 
Specifically, we consider an OOD setting where test images are related to scenes not observed during training. 
We first identify a set of scene-related objects and restrict all training images to only those that do not contain these objects.
For example, in the \textit{Indoor} OOD setting, none of the training images should contain \textit{sofa}, \textit{bed} or any of the other objects that we have identified to be related to \textit{Indoor} scenes.
To build fully-supervised VQA models for comparison, we consider (1) \textbf{BUTD}~\cite{DBLP:conf/cvpr/00010BT0GZ18}, a classic two-stream VQA models, (2) traditional \textbf{NMNs}~\cite{DBLP:conf/cvpr/AndreasRDK16}, and (3) finetuned pre-trained vision-language models, including \textbf{VilBert}~\cite{DBLP:conf/nips/LuBPL19}, \textbf{VisualBert}~\cite{DBLP:journals/corr/abs-1908-03557} and \textbf{ALBEF}~\cite{DBLP:conf/nips/LiSGJXH21}.

The results are shown in Table~\ref{tab:gqa_ood}.
We can see from the table that for those supervised VQA models, when they are trained on images with different scenes, their performance on the target domain is clearly lower than our Mod-Zero-VQA.
Furthermore, our Mod-Zero-VQA method achieves steady performance across different scenes, whereas the supervised VQA models give fluctuated performance in different scenes.
This demonstrates the robustness of our proposed method.

\begin{table}[t]
\centering
\begin{small}
  \begin{tabular}{l|ccc}
    \toprule
    \textbf{Method} & \textbf{Indoor} & \textbf{Food} & \textbf{Street}\\
    \midrule
    BUTD & 39.27&32.28  &35.96 \\
    NMNs &39.45 &32.47 &36.05 \\
    VilBert &39.87 & 32.12&36.68 \\
    VisualBert &41.14&33.47&38.51\\
    ALBEF &45.55&38.87&41.60\\
    \midrule
    Mod-Zero-VQA &48.86 &47.80 &49.54 \\
    \midrule
  \end{tabular}
\end{small}
\caption{Comparison between our Mod-Zero-VQA method and fully-supervised VQA models under the out-of-domain setting.}
  \label{tab:gqa_ood}
\end{table}

\subsection{Case Study}
\begin{figure}[t] 
	\centering
	\setlength{\tabcolsep}{0pt} 
	\renewcommand{\arraystretch}{0} 
	\begin{tabular}{ccc}
		\includegraphics[scale = 0.21]{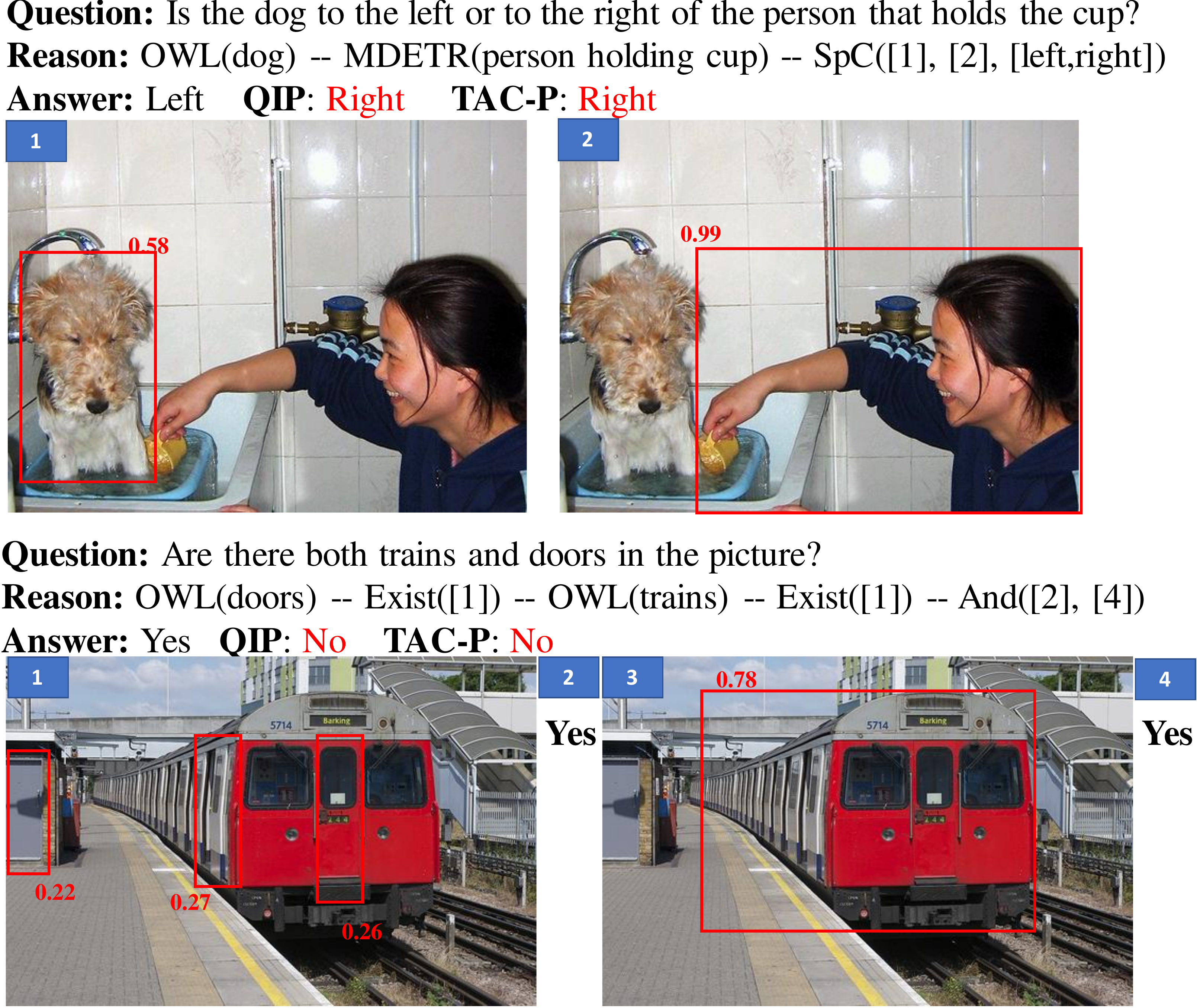} 
	\end{tabular}
	\caption{Visualization of intermediate outputs from reasoning steps of the Mod-Zero-VQA model.}
	\label{fig:case}
\end{figure}

As a case study, we visualize the outputs of the reasoning steps from the proposed method and compare the prediction of the proposed method with those of QIP and TAC-P, which also leverage CLIP as the answer generator. 
We show two example questions and the outputs in Figure~\ref{fig:case}.
Both questions require multiple reasoning steps.

We can see that our method gives the correct predictions while the two other methods answer wrongly. 
We can also see that by decomposing the questions, our method assigns each sub reasoning task to a pre-trained model capable of the task~(i.e., MDETR for reference expression localization and OWL for object detection). 
With question decomposition, we can also better pinpoint the weaknesses of pre-trained models and insert human knowledge by defining simple but general heuristics (e.g., adding spatial heuristics to zero-shot \texttt{CLIP} and defining logical operations). 
More examples with visualization are provided in Appendix~\ref{sec:visualization}.

\section{Related Work}
\label{sec:related}

\subsection{Visual question answering}
Although great progress has been made in the supervised VQA setting~\cite{DBLP:journals/corr/abs-1908-03557,DBLP:conf/nips/LuBPL19,DBLP:conf/icml/0001LXH22,DBLP:conf/nips/LiSGJXH21}, few studies have explored the zero-shot VQA setting.
One line of work converts VQA to text-based QA so that language models (LMs) can be applied. 
Some of them require auxiliary training though not with VQA data~\cite{DBLP:conf/acl/DaiHSJLF22,DBLP:conf/acl/Jin0SC022,DBLP:conf/nips/TsimpoukelliMCE21}. 
Some suffer from insufficient visual details~\cite{DBLP:conf/aaai/YangGW0L0W22} or laborious generation of irrelevant captions~\cite{DBLP:journals/corr/abs-2210-08773}. 
Others~\cite{DBLP:conf/iclr/ShenLTBRCYK22,DBLP:conf/acl/0002000W22} convert VQA to multimodal matching and leverage CLIP~\cite{DBLP:conf/icml/RadfordKHRGASAM21}. 
However, CLIP is limited when compositional reasoning and spatial reasoning are required~\cite{DBLP:journals/corr/abs-2204-03162,DBLP:conf/acl/SubramanianMD0022}.
In this work, we propose to decompose questions and propose a modularized zero-shot VQA method by assigning reasoning tasks to proper pre-trained models without any adaptation.

\subsection{Zero-shot applications of pre-trained models}
Models pre-trained on a large corpus have strong zero-shot transferability when performing down-stream tasks whose objectives are similar to the pre-training objectives of these models. 
For instance, GPT-3~\cite{DBLP:conf/nips/BrownMRSKDNSSAA20} is powerful for zero-shot QA by treating the QA as a text generation problem. CLIP~\cite{DBLP:conf/icml/RadfordKHRGASAM21} demonstrates good zero-shot image recognition capability by treating the classification task as multimodal matching. 
For multimodal QA tasks, LMs can be applied once information from other modalities are translated to tokens LMs understand~\cite{DBLP:journals/corr/abs-2210-08773,DBLP:journals/corr/abs-2206-08155}.
In our work, we decompose VQA questions into sub-reasoning tasks and assign sub-tasks to corresponding pre-trained models whose pre-training objectives match the sub-tasks.

\section{Conclusion and Future Work}
\label{sec:conclusion}
In this work, we propose a modularized zero-shot VQA method, motivated by the idea of Neural Module Network (NMN). 
Instead of training modules in NMN with VQA data, we decompose questions into reasoning tasks explicitly, leverage pre-trained models and assign proper reasoning tasks to them. Experiments show that our model is powerful on questions requiring multi-step reasoning and applicable for real-world VQA. Besides, the proposed model is highly interpretable, which helps to pinpoint weaknesses of a VQA system, making it easier to improve a system. Our model highlights a future direction of leveraging pre-trained models for other complicated tasks requiring multiple reasoning capabilities.

\section*{Limitations}
\label{sec:limitation}
In this section, we discuss few limitations of the proposed method and point out future directions to improve the model. 
First, our method needs to decompose questions into a symbolic representation, but such representations are hard for humans to comprehend, and therefore this decomposition mechanism is hard to be trained with human annotation.
A promising direction is to leverage pre-trained language models such as ChatGPT~\footnote{https://openai.com/blog/chatgpt/} to automate this decomposition step, leveraging ChatGPT's internal knowledge of decomposing a complex question into sub-questions. 
Second, the execution of the zero-shot NMNs is conducted in a deterministic manner, leading to high risks of error propagation if the reasoning chain gets longer. 
In the future, we can consider a softer way of reasoning over the image with pre-trained models.

\section*{Acknowledgement}

This research was supported by the SMU-A*STAR Joint Lab in Social and Human-Centered Computing (Grant No. SAJL-2022-HAS002).

\bibliography{anthology}
\bibliographystyle{acl_natbib}
\clearpage

\appendix

\section{Modules in VQA}
\label{sec:module-vqa}

We summarize all modules in traditional NMNs for VQA~\cite{DBLP:conf/iccv/HuARDS17,DBLP:conf/iclr/GuptaLR0020,DBLP:conf/wacv/ChenGLCW021} in Table~\ref{tab:nmn-module}.

\begin{table*}[ht]
  \begin{tabular}{llp{6.5cm}} 
    \toprule
    \textbf{Module}  & \textbf{Output}  & \textbf{Functionality}\\
    \texttt{Find}($\mathbf{V}$, $\mathbf{g}_{\text{OBJ}}$) & Att. & Locate a certain object (OBJ) in the image\\
    \texttt{Relocate}($\bm{\alpha}$,$\mathbf{V}$,$\mathbf{g}_{\text{RELA}}$) &Att.&   Transit attention from previous attention map $\bm{\alpha}$ according to the relation (RELA)\\
    \texttt{Filter}($\bm{\alpha}$,$\mathbf{V}$,$\mathbf{g}_{\text{CONDI}}$) & Att. & Highlight objects that are attended by previous attention map $\bm{\alpha}$ and satisfy the condition (CONDI)\\
    \texttt{Choose}($\bm{\alpha}_1$,$\bm{\alpha}_2$,$\mathbf{V}$,$\mathbf{g}_{\text{RELA}^1}$,$\mathbf{g}_{\text{RELA}^2}$) & Ans.& Choose the relation  from RELA$^1$ and RELA$^2$ between highlighted regions of two attention maps\\
    \texttt{Query}($\bm{\alpha}$,$\mathbf{V}$,$\mathbf{g}_{\text{QUERY}}$) & Ans.& Generate a final answer given the attention map, image representation and item to query (QUERY)\\
    \texttt{Count}($\bm{\alpha}$) & Ans. &Outputs a number given the attention map of the image \\
     \texttt{Exist}($\bm{\alpha}$) & Ans. & Output a binary answer (\textit{yes/no}) given the attention map of the image \\
    \texttt{And}($\bm{\alpha}_1$,$\bm{\alpha}_2$) & Ans.& Generate a binary answer (\textit{yes/no}) given the two attention maps \\
    \texttt{Or}($\bm{\alpha}_1$,$\bm{\alpha}_2$) & Ans. &  Generate a binary answer (\textit{yes/no}) given the two attention maps\\
    \bottomrule
\end{tabular}
\centering
\caption{The full list of modules in traditional NMNs. $\mathbf{g}_{[\cdot]}$ is the word embedding for the words in $[\cdot]$.
}
\label{tab:nmn-module}
\end{table*}

\section{Logical Operations}
\label{sec:logical-op}
In this section we describe the logical modules $\texttt{And}$ and $\texttt{Or}$. Both of them receive outputs from two zero-shot $\texttt{Exist}$ modules. For the \texttt{And} module, if both outputs are \textit{yes}, it outputs \textit{yes}; otherwise, it outputs \textit{no}. For the \texttt{Or} module, if both outputs are \textit{no}, it outputs \textit{no}; otherwise, it outputs \textit{yes}. The logical operators are deterministic. 

\section{Existence Questions}
\label{sec:existence-ques}

\begin{figure}[t] 
	\centering
	\setlength{\tabcolsep}{0pt} 
	\renewcommand{\arraystretch}{0} 
	\begin{tabular}{ccc}
		\includegraphics[scale = 0.2]{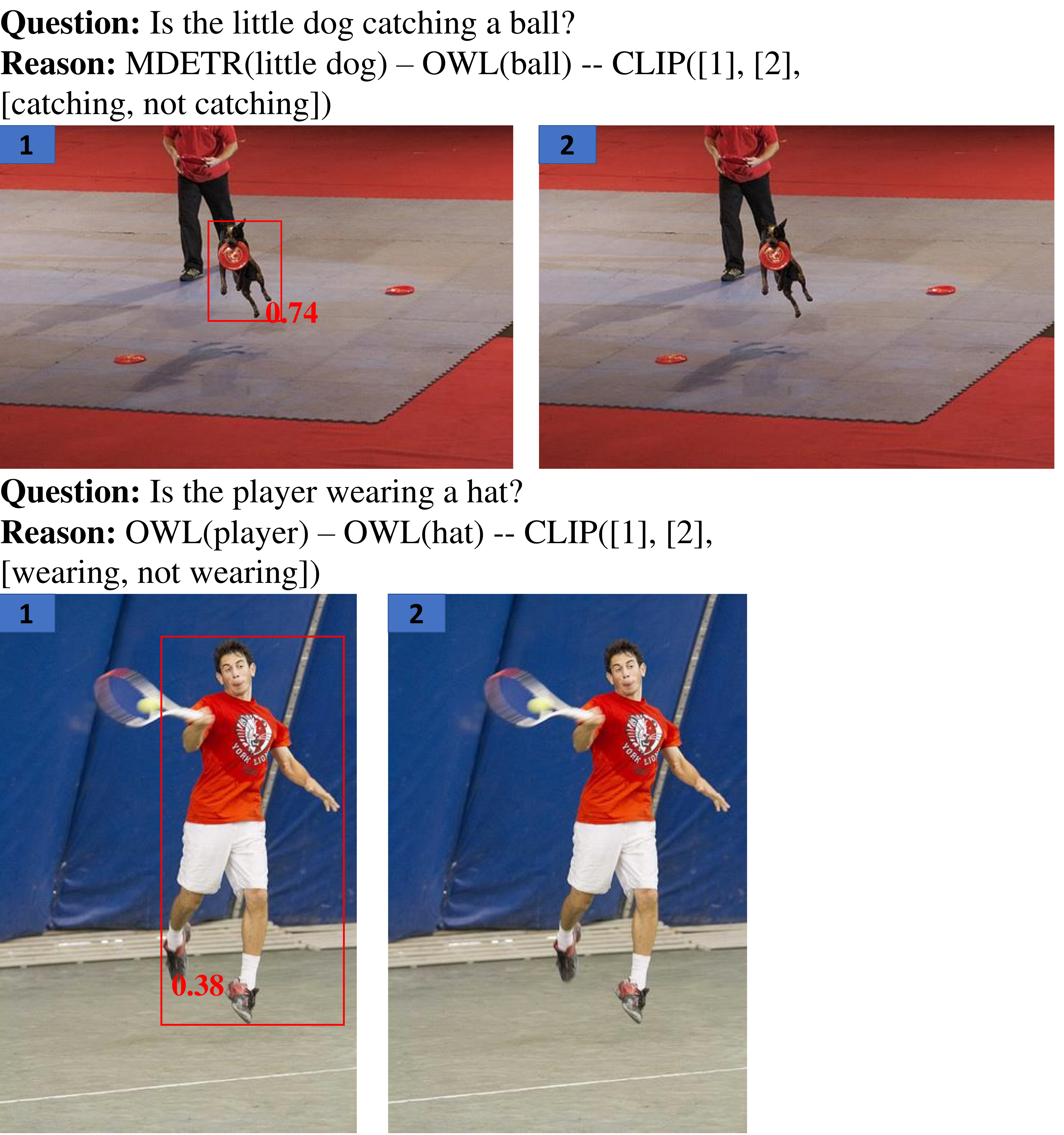} 
	\end{tabular}
	\caption{Visualization of existence-related questions where mentioned objects in the questions do not exist.}
	\label{fig:exist-no}
\end{figure}

\begin{figure}[t] 
	\centering
	\setlength{\tabcolsep}{0pt} 
	\renewcommand{\arraystretch}{0} 
	\begin{tabular}{ccc}
		\includegraphics[scale = 0.2]{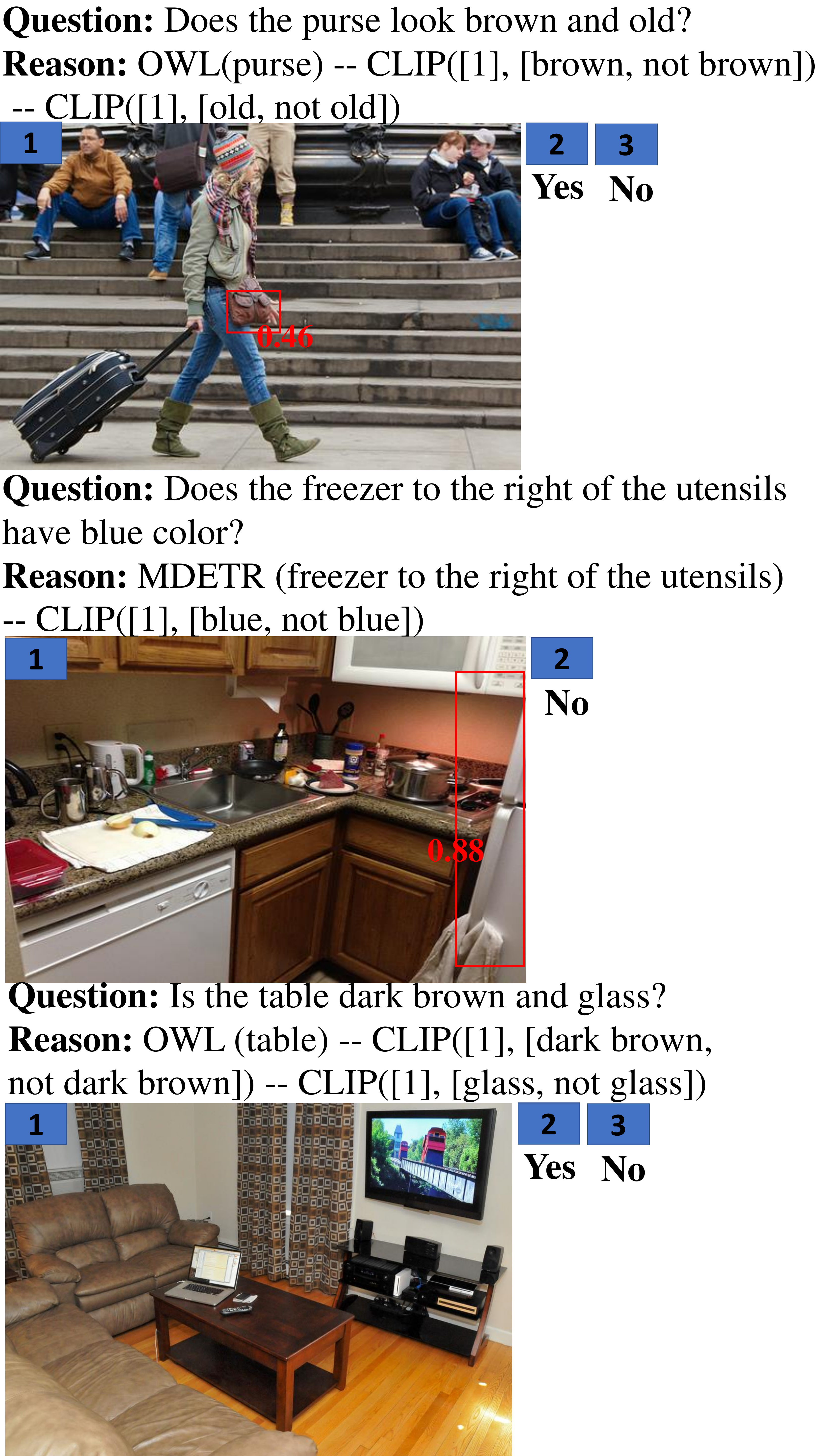} 
	\end{tabular}
	\caption{Visualization of questions asking about existence of attributes.}
	\label{fig:exist-attr}
\end{figure}

\begin{figure}[t] 
	\centering
	\setlength{\tabcolsep}{0pt} 
	\renewcommand{\arraystretch}{0} 
	\begin{tabular}{ccc}
		\includegraphics[scale = 0.2]{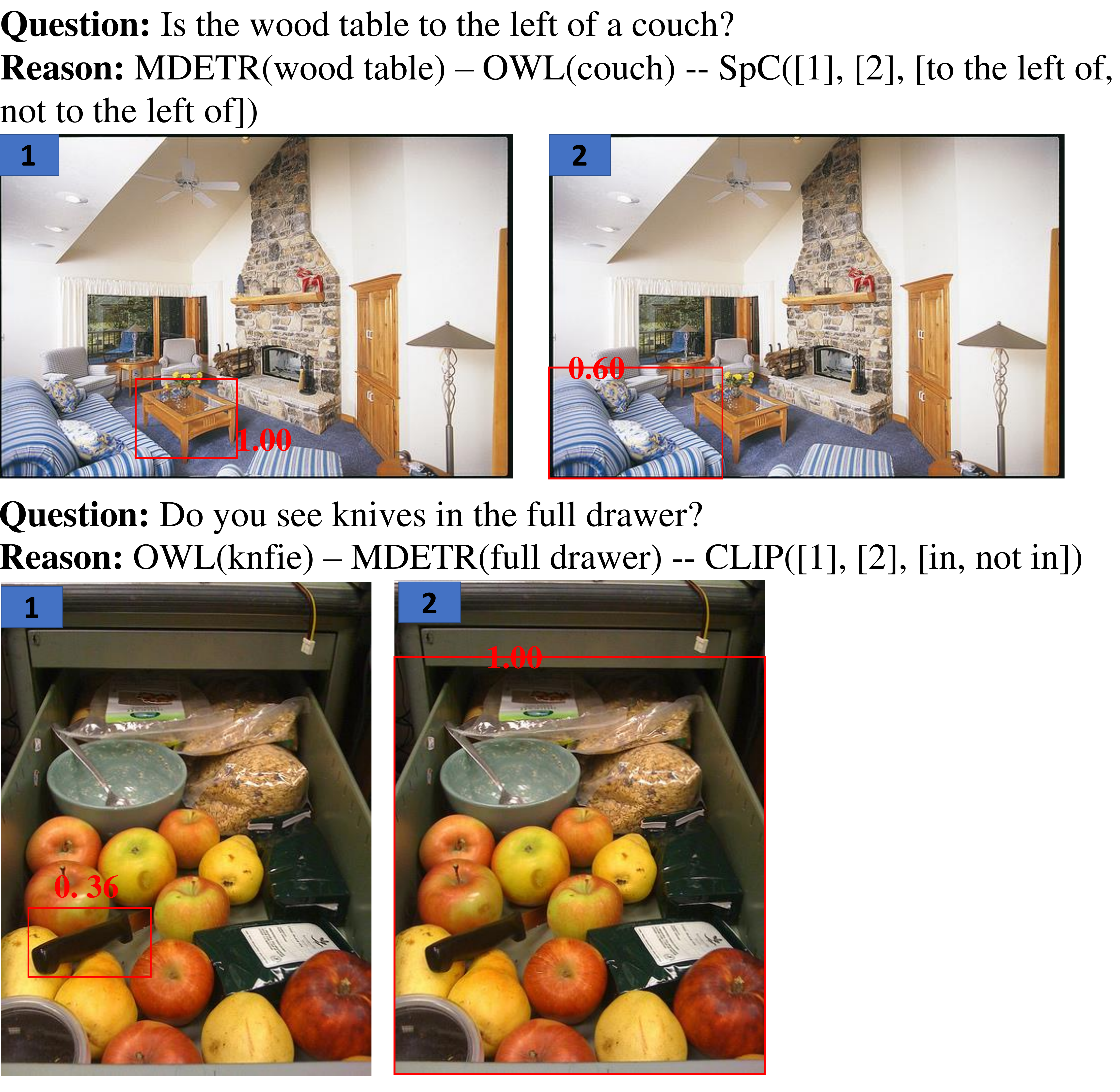} 
	\end{tabular}
	\caption{Visualization of questions asking about the existence of relations.}
	\label{fig:exist-rela}
\end{figure}

As mentioned briefly in Section~\ref{subsec:mapping}, for questions verifying the existence of something, according to the NMN layout, we classify these questions into three types: verifying existence of objects, of attributes, and of relations. For the verification of object existence, we directly apply the zero-shot \texttt{OWL}. For both attribute and relation verification questions, we first make sure all objects mentioned in the question exist with the help of \texttt{OWL}. If any mentioned objects do not exist, the predicted answer will be \textit{No}, as illustrated in Figure~\ref{fig:exist-no}. If the objects exist, we leverage either zero-shot \texttt{OWL} or \texttt{MDETR} to locate at objects of interests and verify the attributes and relations, with the utilization of the CLIP module. Examples are provided in Figure~\ref{fig:exist-attr} (for attribute verification) and Figure~\ref{fig:exist-rela} (for relation verification). We use CLIP for binary matching to select whether the attribute/relation exists. When multiple attributes/relations are to be verified, only when all attributes/relations exist will the predicted answer be \textit{Yes}; otherwise, the prediction is \textit{No}. For instance, the third example in Figure~\ref{fig:exist-attr} has a dark brown table, but the table is not glass, so the third step outputs \textit{no}. The final predicted answer to the question is therefore \textit{no}.

\section{Detailed Implementation for Spatial Heuristics}
\label{sec:sp-heur-detail}
In this section, we give the mathematical definitions of the spatial heuristics. The input bounding box is denoted as $\mathbf{b}=(x,y,w,h)$, representing the relative position and relative size of the object in the VQA image.

\noindent\textbf{\texttt{Spatial Determine (SpD)}} receives an object bounding box and determines which position in the original image the object is at. The position candidates $\mathcal{P}$ are generated according to the question. When the question is asking for the horizontal position of the object, $\mathcal{P} = \{\text{left, right}\}$; When the question is asking for the vertical position of the object, $\mathcal{P} = \{\text{top, bottom}\}$. 
The \texttt{SpD} module is implemented as:
\begin{equation}
    \texttt{SpD}(\mathbf{b}, \mathcal{P}) = \left\{
    \begin{aligned}
         &  \text{left}, \quad \text{if}\quad  x<0.5\\
         &  \text{right}, \quad \text{else}
    \end{aligned}
    \right .
\end{equation}
when $\mathcal{P} = \{\text{left, right}\}$. When $\mathcal{P} = \{\text{top, bottom}\}$, the spatial heuristic is derived as:
\begin{equation}
    \texttt{SpD}(\mathbf{b}, \mathcal{P}) = \left\{
    \begin{aligned}
         &  \text{top}, \quad \text{if}\quad  y<0.5\\
         &  \text{bottom}, \quad \text{else}
    \end{aligned}
    \right .
\end{equation}
The SpD heuristic will be used in the \texttt{Query} module when asking about either horizontal or vertical position.

\noindent\textbf{Spatial Chooser (SpC)} receives two bounding boxes of objects $\mathbf{b}_1, \mathbf{b}_2$ and aims to choose their spatial relations from the relation candidates in $\mathcal{C}$ ($\mathbf{b}_1$ is \textit{RELA} $\mathbf{b}_2$). For instance, when $\mathcal{C}= \{\text{to the left of},\text{to the right of}\}$:
\begin{equation}
    \texttt{SpC}(\mathbf{b}_1, \mathbf{b}_2, \mathcal{C}) = \left\{
    \begin{aligned}
         &  \text{left}, \text{if}\quad  x_{1}<x_{2}\\
         &  \text{right}, \quad \text{else}
    \end{aligned}
    \right .
\end{equation}
When $\mathcal{C}= \{\text{above},\text{beneath}\}$:
\begin{equation}
    \texttt{SpC}(\mathbf{b}_1, \mathbf{b}_2, \mathcal{C}) = \left\{
    \begin{aligned}
         &  \text{above}, \text{if}\quad  y_{1}<y_{2}\\
         &  \text{beneath}, \quad \text{else}
    \end{aligned}
    \right .
\end{equation}
The SpC rule will be applied to the \textit{Choose} type of questions
if the choices of relations fall into the sets below:
 [\{to the left of\},\{to the right of\}] and 
[\{above, on top of\},\{under, below, beneath, underneath\}]

\section{Dataset Statistics}
\label{sec:dataset-st}

In Table~\ref{tab:dataset-dis}, we provide statistics of the GQA and the VQA dataset. Following~\cite{DBLP:conf/acl/0002000W22,DBLP:conf/nips/TsimpoukelliMCE21}, we use the validation split for testing. Specifically, we report \textit{soft vqa scores} as there may be multiple possible answers to a question similar to previous works. ~\cite{DBLP:conf/acl/0002000W22,DBLP:conf/nips/TsimpoukelliMCE21,DBLP:conf/cvpr/00010BT0GZ18,DBLP:conf/emnlp/FukuiPYRDR16}.

\begin{table}[t]
\centering
\begin{small}
  \begin{tabular}{c|cc|cc}
    \toprule
    \textbf{Dataset}& \multicolumn{2}{c|}{\textbf{Train}} & \multicolumn{2}{c}{\textbf{Val}}\\
     & \textbf{\# Ques.} & \textbf{\# Img.} & \textbf{\# Ques.} & \textbf{\# Img.} \\
    \midrule 
    GQA &943,000 &72,140  &132,062  &10,234 \\
    VQA & 443,757&82,783  & 214,354 &  40,504\\
    \bottomrule
\hline
\end{tabular}
\end{small}
\caption{Statistical distributions of the GQA and the VQA dataset.}
\label{tab:dataset-dis}
\end{table}

\section{Layout Generation}
\label{sec:layout-gen}
The layout generation can be accomplished either with syntatic parser or a pre-trained sequence-to-sequence layout generator. On the VQA dataset, we follow~\cite{DBLP:conf/cvpr/AndreasRDK16,DBLP:conf/iccv/HuARDS17} to parse questions with Stanza\footnote{https://github.com/stanfordnlp/stanza} and transform the parsed tree into reasoning graphs where each node is a pre-defined module with rules most similar to~\cite{DBLP:conf/iccv/HuARDS17}. The graphs are converted to module sequences with the post-order traversal. The linearlized module sequence is used as the layout. On GQA dataset, we leverage layouts generated by the pre-trained sequence-to-sequence layout generator from~\cite{DBLP:conf/wacv/ChenGLCW021}. The generator adopts a coarse-to-fine two-stage generation paradigm as in~\cite{DBLP:conf/acl/LapataD18} to encode questions and decode the sequence of module names and module inputs in two stages.

\section{Answer Filtering}
\label{sec:ans-gen}
Basically, we follow~\cite{DBLP:conf/acl/0002000W22} to narrow down the set of possible answer candidates with the language model T5~\cite{DBLP:journals/jmlr/RaffelSRLNMZLL20}. For the VQA dataset, we directly leverage the published generated candidate answers for each question from the paper~\cite{DBLP:conf/acl/0002000W22}. For the GQA dataset, the \textit{Verify} and \textit{Logical} type questions have binary answers \textit{yes/no}. For the \textit{Compare} and \textit{Choose}, candidate answers are available in the generated layouts. For the \textit{Query} type of questions, we first convert questions into masked templates with a rule-based converter~\cite{DBLP:journals/corr/abs-1809-02922}. T5 is applied to retrieve the masked word, which filters out irrelevant answers in the answer vocabulary according to contexts.

\begin{table*}[t]
\centering
\begin{small}
  \begin{tabular}{l|cc|cccc}
    \toprule
    \textbf{Backbone} & \multicolumn{2}{c|}{\textbf{Yes/No}} & \multicolumn{4}{c}{\textbf{Other}}\\
     & \textbf{Verify} & \textbf{Logical}& \textbf{Choose} & \textbf{Compare}&\textbf{Query}&\textbf{Overall}\\
    \midrule
    ViT-B/16 &69.63&68.63&75.87&48.59&26.36&47.28 \\
    \text{Res}50$\times$ 16&68.51&68.71&75.78&41.84&25.69& 46.49\\
    ALBEF &68.08&69.99&75.93&48.40&29.38&48.68\\
    \midrule
\end{tabular}
\end{small}
\caption{Performance of the proposed model with different models for multimodal matching regarding different question types.}
  \label{tab:detailed_gqa}
\end{table*}

\label{sec:visualization}
\begin{figure*}[t] 
	\centering
	\setlength{\tabcolsep}{0pt} 
	\renewcommand{\arraystretch}{0} 
	\begin{tabular}{ccc}
		\includegraphics[scale = 0.2]{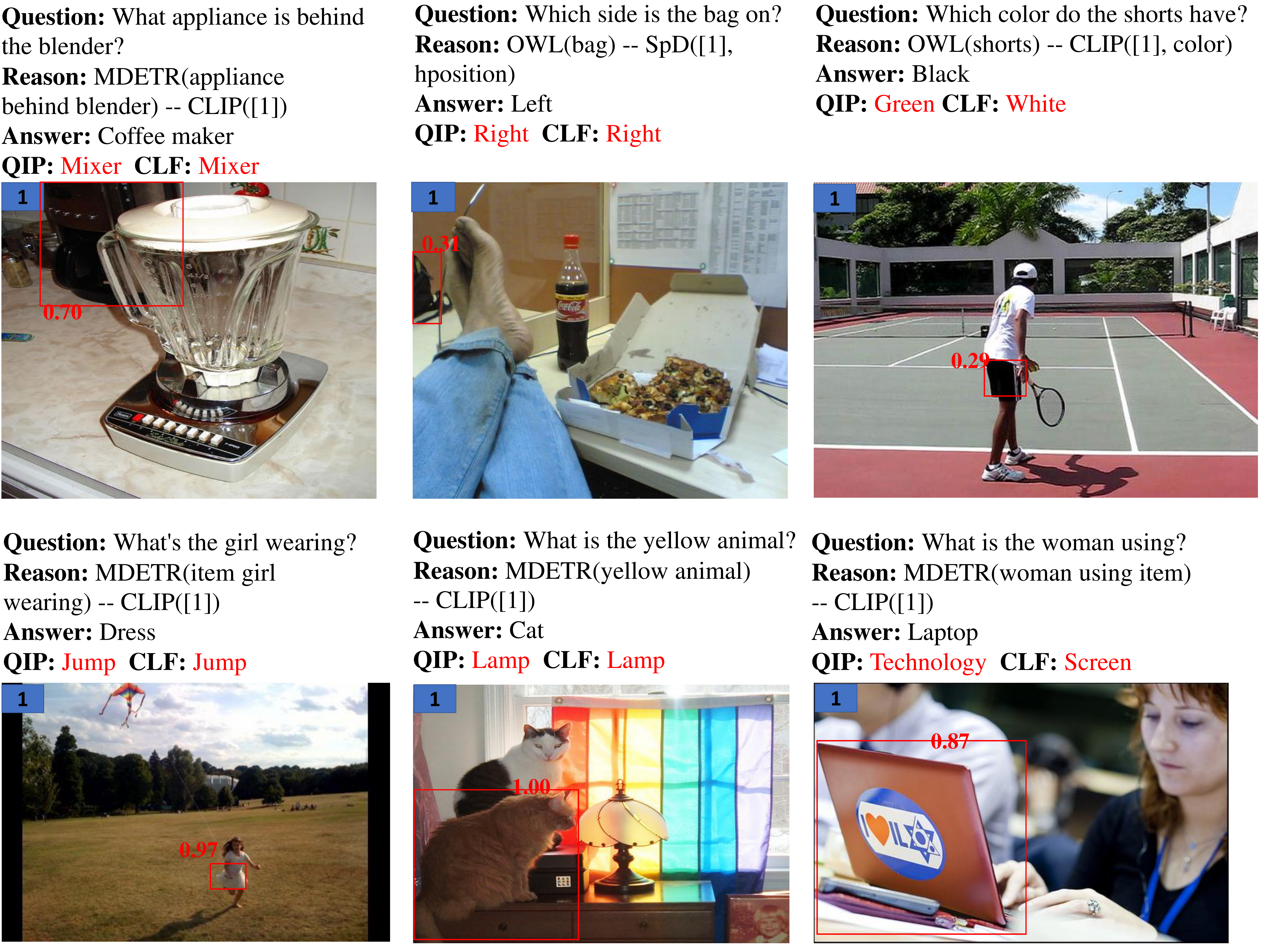} 
	\end{tabular}
	\caption{Visualization of VQA examples with short reasoning chains.}
	\label{fig:case-single}
\end{figure*}

\begin{figure*}[t] 
	\centering
	\setlength{\tabcolsep}{0pt} 
	\renewcommand{\arraystretch}{0} 
	\begin{tabular}{ccc}
		\includegraphics[scale = 0.2]{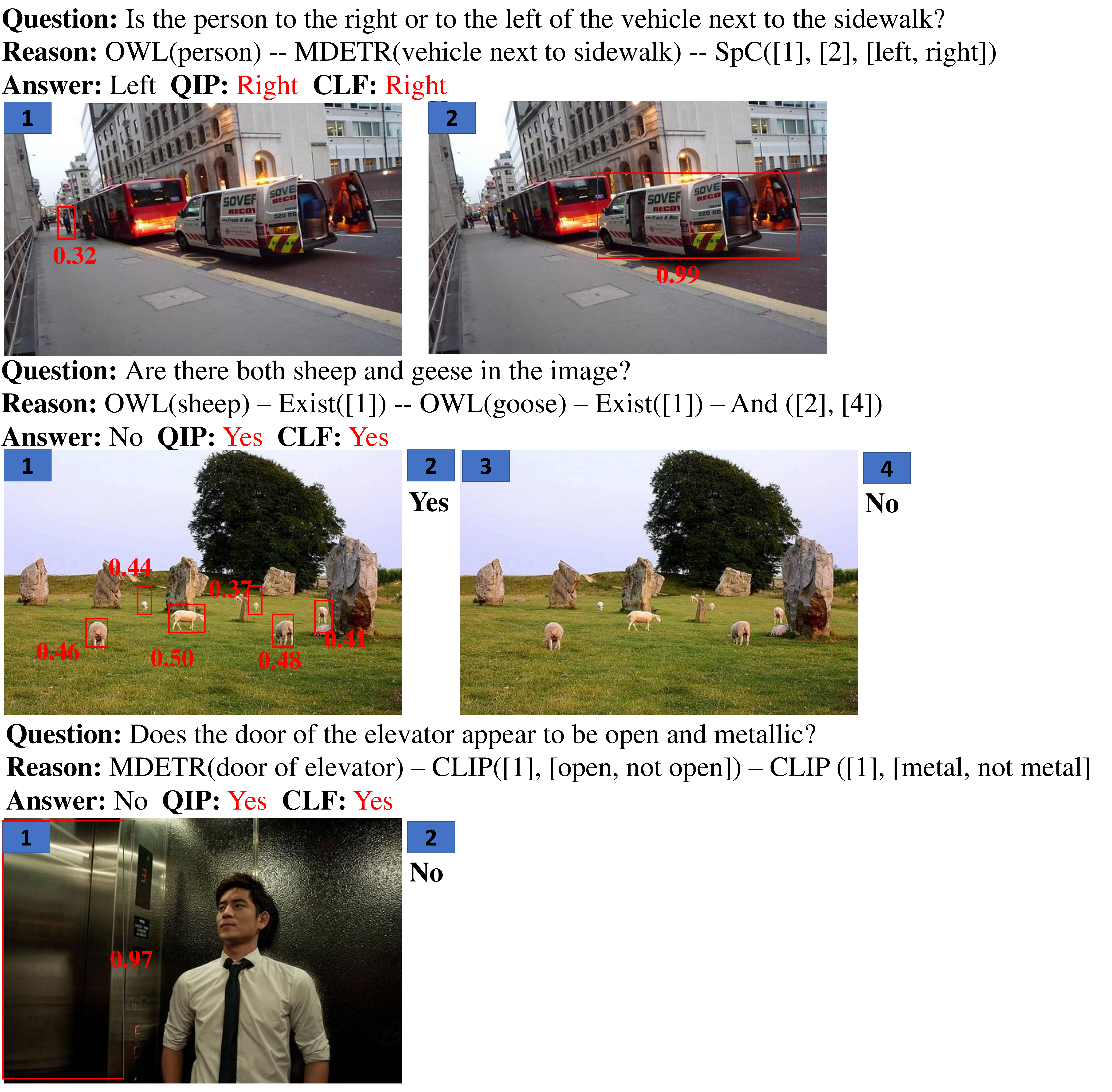} 
	\end{tabular}
	\caption{Visualization of VQA examples with long reasoning chains.}
	\label{fig:case-multi}
\end{figure*}

\label{sec:exp-setup}
\begin{figure*}[t] 
	\centering
	\setlength{\tabcolsep}{0pt} 
	\renewcommand{\arraystretch}{0} 
	\begin{tabular}{ccc}
		\includegraphics[scale = 0.27]{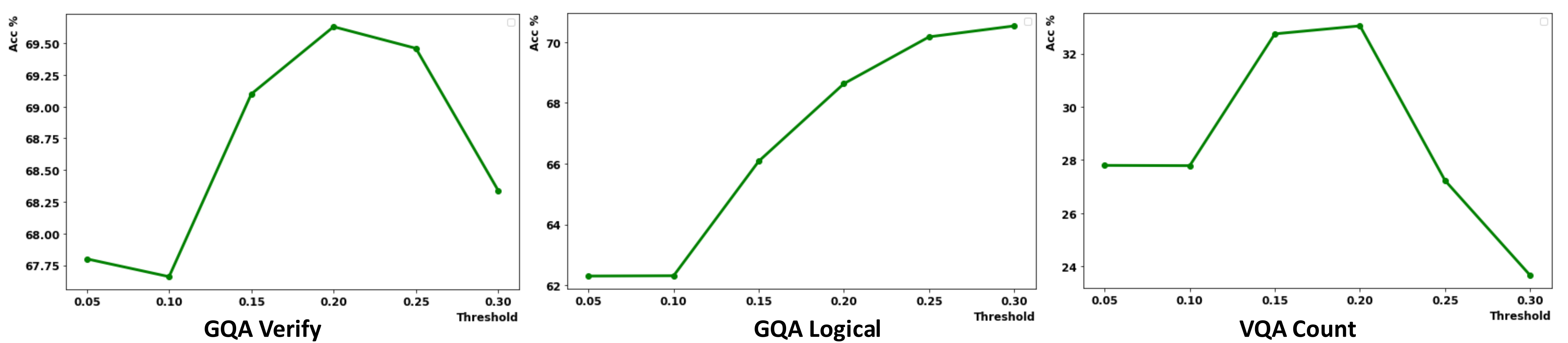} 
	\end{tabular}
	\caption{Performances of the zero-shot VQA model regarding to different thresholds of confident scores in the OWL model.}
	\label{fig:hyper}
\end{figure*}

\section{Detailed Results}
\label{sec:detailed-results}
We provide the detailed results for replacing CLIP with ALBEF (discussed in Section~\ref{sec:ablation}) in Table~\ref{tab:detailed_gqa} considering different types of questions.

\section{Visualization of Zero-shot NMNs}
In this section, we provide more visualization examples which the zero-shot NMNs answers correctly while the baselines (QIP and TAC-P) fail. In Figure~\ref{fig:case-single}, we show examples with short reasoning chain, specifically, only two-step in Mod-Zero-VQA. According to the results, we observe that each intermediate step gives interpretable outputs. 
By question decomposition and leveraging pre-trained models, our model can focus on relevant regions of the image (e.g., the first and third example in the first row of Figure~\ref{fig:case-single}) so that eliminating noise from backgrounds. Without filtering irrelevant information in the image, baselines pay attention to dominant objects in the image, leading to wrong predictions (e.g., the third example in the first row which QIP and TAC-P seems to focus on the ground and the T-shirt when answering the question). 
In Figure~\ref{fig:case-multi}, we visualize questions with relatively longer reasoning chains. These compositional questions usually call several reasoning capabilities, making it hard for pre-trained VL models to deal with~\cite{DBLP:journals/corr/abs-2204-03162}. With question decomposition, each pre-trained model takes a sub reasoning task, easing the burden from answering a complicated question.

According to the visualization, we also find a frequent error case resulting from the wrongly-generated NMN layout. The coarse-to-fine two stage generation suffers from the issue of early stopping that the generated arguments is incomplete. For instance, the ground-truth step should be \texttt{Find}(\textit{coffee table}) while the generated result is \texttt{Find}(\textit{coffee}).

\section{Out-of-Distribution Setting Construction}
\label{sec:ood-const}
We consider an Out-of-Domain Generalization (OOD) setting, where test images are related to scenes (i.e., \textit{Indoor}, \textit{Food} and \textit{Street}) not observed during training. For the \textit{Indoor} scene, we directly leverage the annotation from Visual Genome~\cite{DBLP:journals/ijcv/KrishnaZGJHKCKL17}, where images are classified as indoors and outdoors. For the other two settings, we filter out training images containing those scene-specific objects and make sure a certain protion of objects in the testing images are about those objects (in other words, testing images are related to the scene).
Below, we provide the lists of scene specific objects in the \textit{Food} and \textit{Street} scene.

\noindent \textbf{Food: }
plate, banana, table, food, pizza, donut, fork, bowl, cheese, napkin, glass, cake, tomato, bread, apple, carrot, knife, broccoli, vegetable, fruit, cup, sauce, orange, spoon, meat, pepper, crust, onion, sandwich, home plate, topping, catcher, tray, lettuce, container, dish, bottle, batter, umpire, frosting, hot dog, egg, chicken, bat, box, mask, paper, mushroom, mug, pitcher, dispenser, liquid, label, bacon, tablecloth, nut, leaf, utensil, salad, hand, crumb, lemon, basket, mound, card, helmet, strawberry, lid, pan, seed, chair, menu, jar, player, sausage, icing, juice, shirt, spinach, sprinkle, dugout, counter, bag, flower, berry, goat, sailboat, uniform, steering wheel, glove, heel, pastry, bubble, finger, sugar, beer, oven, heart, dessert, herb

\noindent \textbf{Street: }
car, sign, building, pole, letter, tree, tire, road, wheel, sidewalk, bus, train, street, number, door, sky, bike, windshield, truck, street light, motorcycle, leaf, traffic light, roof, ground, post, license plate, arrow, vehicle, fence, cloud, word, grass, wire, van, bicycle, gravel, bush, platform, fire hydrant, house, seat, flag, bag, pavement, step, graffiti, sticker, logo, paint, luggage, cone, chain, pipe, helmet, bridge, balcony, parking lot, jacket, plant, stop sign, train car, umbrella, taxi, lamp, box, crosswalk, flower, bench, brick, store, trash can, clock, gate, station, jean, grill, suv, driver, hook, pant, trash, tower, city, stair, rock, coat, rose, chimney, trailer, american flag, entrance

\section{Experiment Settings}
In this section, we discuss the experiment settings regarding to the size of models, method of choosing hyper-parameters and the used software packages and versions.

\noindent\textbf{Model Size:} We provide the number of parameters of different models in Table~\ref{tab:num-params}. Our model includes the OWL model, the MDETR model, the CLIP$_{\text{ViT-B/16}}$ and the T5 model for answer filtering. It consists of $1,521$M parameters, of which the T5 model takes $770$M parameters, the OWL model takes $583$M parameters, the MDETR model takes $170$M parameters and the CLIP model takes $151$M parameters. After pre-processing object detection and answer filtering, it takes $6$G GPU memory for inference.
\begin{table}[h]
\centering
\begin{small}
  \begin{tabular}{l|l}
    \toprule
    \textbf{Method} & \textbf{\# Params (M)}  \\
    \midrule
    VL-T5$_{\text{no-vqa}}$  &288\\
   FEWVLM$_{\text{base}}$&288 \\ 
    FEWVLM$_{\text{large}}$ & 804\\
    VLKD$_{\text{ViT-L/14}}$ & 713\\
    BNP-VQA$_{6\text{M}}$ &669 \\
    BNP-VQA$_{11\text{B}}$ &1,576 \\
    Frozen &1,040 \\
    $\text{QIP}_{\text{ViT-B/16}}$ & 151\\
    $\text{TAC-P}_{\text{ViT-B/16}}$ &921 \\
    \hline
    Zero-shot NMNs & 1,521\\
    \midrule
\end{tabular}
\end{small}
\caption{Number of parameters in VQA models.}
  \label{tab:num-params}
\end{table}

\noindent\textbf{Hyper-parameters:}
As we focus on the zero-shot learning setting so that there is no training process. Here we provide hyper-parameters used as thresholds. For the OWL model~\cite{DBLP:journals/corr/abs-2205-06230}, we set the threshold of confident score as $0.2$, which is set empirically, to filter out detected bounding boxes of which the confident scores are too long.
We show test the robustness of the proposed zero-shot VQA model regarding to the hyper-parameter of the threshold and provide experimental results corresponding to the threshold varying from $0.05, ,0.1, 0.15, 0.2, 0.15, 0.3$ in Figure~\ref{fig:hyper}. As proven in Section~\ref{sec:ablation}, the detection result mostly affects binary questions which rely more on object detection results, we here provide results for \textit{Verify} and \textit{Logical} type of questions on GQA. Besides, \textit{Count} type questions also heavily rely on the quality of object detection. According to the results, we observe the zero-shot NMNs achieves relatively stable performances regarding to different thresholds for confident scores on \textit{Verify} type questions, while less stable for the \textit{Logical} and  \textit{Count} type questions. The stability on \textit{Verify} questions depicts the robustness of the detection model. As \textit{Logical} questions combines results from two \textit{Verify} questions, the error may propagate if one predicted answer of the \textit{Verify} question is wrong. An interesting finding is that the performance does not drop as the threshold increases. This may be that answers are biased to \textit{no}. With the increment of thresholds, the model is more likely to answer \textit{no}. \textit{Count} questions are more sensitive to the threshold because lower thresholds lead to the case that uncertain regions to be detected while higher thresholds are more harmful that correctly detected objects will be filtered out. In conclusion, the threshold is important to the quality of detection and setting it from $0.2$ to $0.25$ gives good performances. 
For the MDETR model, we directly follow their published code for detection and set the threshold as $0.7$\footnote{https://colab.research.google.com/drive/11xz5IhwqAqHj9-XAIP17yVIuJsLqeYYJ?usp=sharing}.

\noindent\textbf{Package Version:} We list the software packages used as well as the corresponding versions in Table~\ref{tab:pack-version}.
\begin{table}[h]
\centering
\begin{small}
  \begin{tabular}{c|c}
    \toprule
    \textbf{Package} & \textbf{Version)}  \\
    \midrule
    PyTorch & 1.9.0\\
    Transformers & 4.19.2\\
    Stanza & 1.4.0\\
    NLTK &3.2.5 \\
    \midrule
\end{tabular}
\end{small}
\caption{Versions of packages used in our experiments.}
  \label{tab:pack-version}
\end{table}

\end{document}